\documentclass[11pt,a4paper]{article}
\usepackage[hyperref]{emnlp2020}
\usepackage{times}
\usepackage{latexsym}

\usepackage{microtype}
\usepackage{hyperref}
\usepackage{url}
\newcommand{\ignore}[1]{}

\newcommand{\bert}{\textsc{Bert}\xspace}

\newcommand{\gpt}{\textsc{Gpt-2}\xspace}
\newcommand{\ff}{\textsc{Family-Feud}\xspace}

\usepackage{amsthm}
\newtheorem*{remark}{Remark}

\usepackage[multiple]{footmisc}

\usepackage{pslatex}
\usepackage{latexsym}
\usepackage{amsfonts}
\usepackage{mathtools}
\usepackage{graphicx}
\usepackage{amsmath}
\usepackage{xspace}
\usepackage{tikz-dependency}
\usepackage{balance}
\usepackage{enumitem}
\usepackage{mathtools}
\usepackage{algorithm}
\usepackage{algorithmicx}
\usepackage{eqparbox}

\usepackage{algcompatible}

\usepackage{xcolor}
\definecolor{beaublue}{rgb}{0.74, 0.83, 0.9}
\definecolor{peachorange}{rgb}{1.0, 0.8, 0.6}
\definecolor{lavendergray}{rgb}{0.77, 0.76, 0.82}
\definecolor{lemonchiffon}{rgb}{1.0, 0.98, 0.8}
\definecolor{mintcream}{rgb}{0.93, 1.0, 0.94}

\usepackage{footnote}
\makesavenoteenv{tabular}

\usepackage{varwidth}
\usepackage{pifont}
\usepackage{float}
\usepackage{framed} 
\usepackage{verbatim}
\usepackage{pgfplots}
\usepackage[rightcaption]{sidecap}
\usepackage{subfig}
\usepackage{wrapfig}
\usepackage{multirow}
\usepackage{soul}
\DeclareSymbolFont{extraup}{U}{zavm}{m}{n}
\DeclareMathSymbol{\varheartsuit}{\mathalpha}{extraup}{86}

\usepackage{xargs} 
\usepackage[colorinlistoftodos,prependcaption,textsize=tiny]{todonotes}

\usepackage{marginnote}
\usepackage{lipsum}

\usepackage{microtype}
\usepackage{booktabs}
\usepackage{rotating}

\definecolor{largest}{HTML}{F8A102}
\definecolor{second}{HTML}{F1DC58}
\definecolor{third}{HTML}{FFCCC9}
\definecolor{small}{HTML}{DAE8FC}

\aclfinalcopy 

\title{ProtoQA: A Question Answering Dataset for \\Prototypical Common-Sense Reasoning}

\author{Michael Boratko\thanks{\quad Equal contribution.} \qquad Xiang Lorraine Li\footnotemark[1] \qquad Tim O'Gorman\footnotemark[1] \qquad Rajarshi Das\footnotemark[1] \\ \qquad \textbf{Dan Le} \qquad \textbf{Andrew McCallum} \\
  College of Information and Computer Sciences \\
  University of Massachusetts Amherst\\
  \texttt{\{mboratko,xiangl,rajarshi,togorman,dhle,mccallum\}@cs.umass.edu}}

\date{}

\begin{document}
\maketitle
\begin{abstract}
Given questions regarding some prototypical situation --- such as \emph{Name something that people usually do before they leave the house for work?} --- a human can easily answer them via acquired experiences. There can be multiple right answers for such questions, with some more common for a situation than others.

This paper introduces a new question answering dataset for training and evaluating common sense reasoning capabilities of artificial intelligence systems in such prototypical situations. The training set is gathered from an existing set of questions played in a long-running international game show -- \ff. The hidden evaluation set is created by gathering answers for each question from 100 crowd-workers. We also propose a generative evaluation task where a model has to output a ranked list of answers, ideally covering all prototypical answers for a question. After presenting multiple competitive baseline models, we find that human performance still exceeds model scores on all evaluation metrics with a meaningful gap, supporting the challenging nature of the task.
\end{abstract}

\section{Introduction}

\label{introduction}
\begin{figure}[t]
    \centering
    \includegraphics[width=\columnwidth]{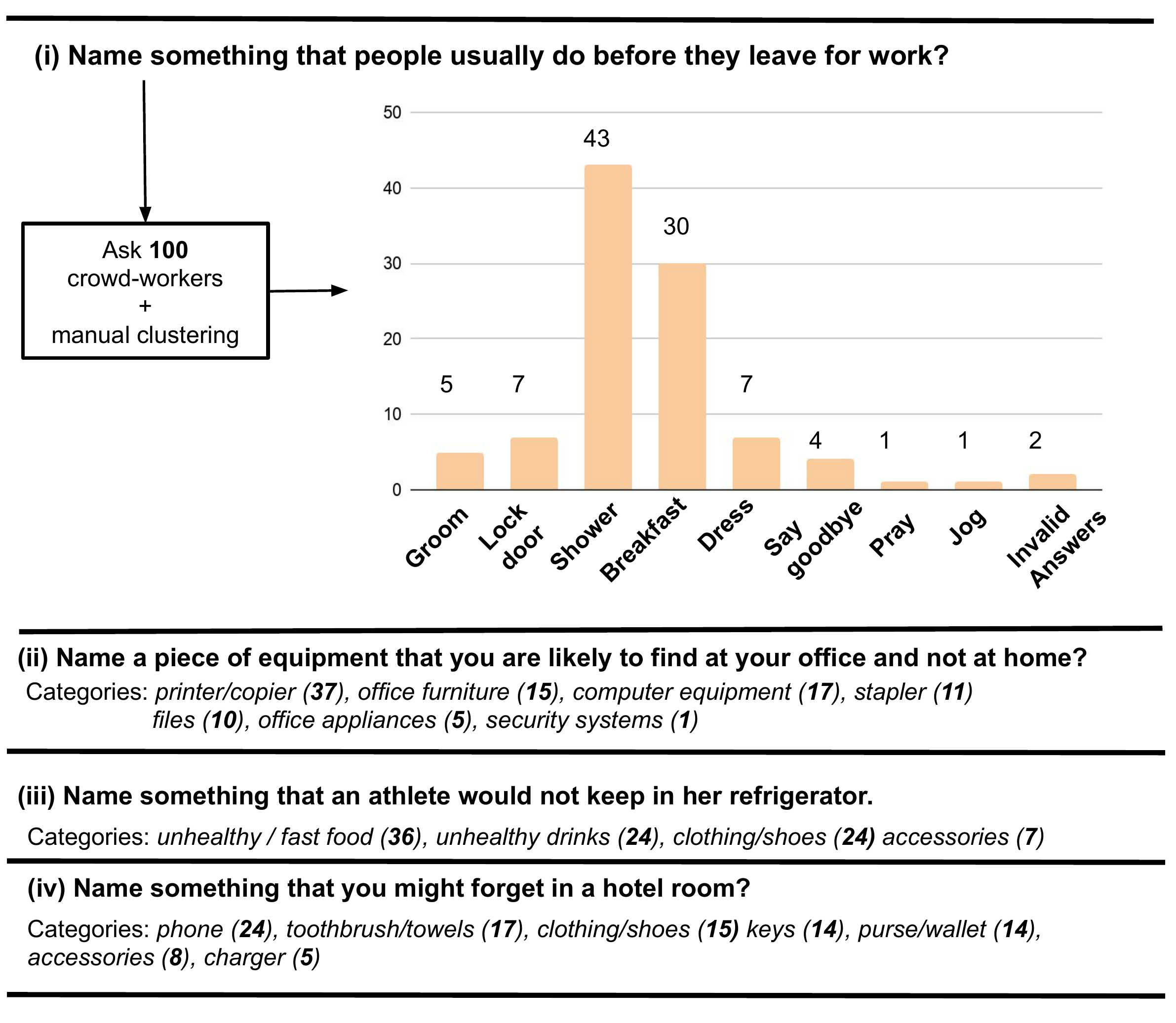}
    \caption{We focus on common-sense reasoning over prototypical situations when there could be many different answers but some are more common than others. Our task is in generative style (\emph{not} multiple-choice format). Answers to a question are crowd-sourced from 100 workers and are then manually clustered into categories. To perform well, a model has to output a ranked list of answers covering multiple categories.}
    \label{fig:intro_fig}
\end{figure}

Humans possess the ability to implicitly reason using a wealth of common background knowledge, much of which is acquired through shared experiences. For example, consider the question in Figure~\ref{fig:intro_fig} --- ``\textit{Name something that people usually do before they leave the house for work.}''. Humans can agree about the details and characteristics of a prototypical event or situation~\citep{schank1975scripts,schank1977scripts} due to commonalities in their shared lived experiences, cultural norms and expectations.  
This rough agreement extends beyond an agreement on a single top response, but can be viewed as a ranked list of plausible answers, 
as demonstrated in Figure~\ref{fig:intro_fig}. 
Such sets of diverse answers represent the nature of common sense knowledge and may be useful in applications such as dialogue systems, where multiple responses are appropriate for a given context~\citep{Zhang2019TaskOrientedDS}.

We present a new question/answer dataset capturing both the plausibility of the answers and the ranking preference of each answer about such prototypical situations inspired by the long-running American game show \ff, which also provides the training data for the task.\footnote{Dataset: \url{https://github.com/iesl/protoqa-data}. \\ Interactive demo: \url{http://protoqa.com}. } The game show is played by prompting participants with queries such as \textit{Name something that people usually do before they leave the house for work} (as shown in Figure~\ref{fig:intro_fig}).
The answers to such questions are provided by 100 randomly selected individuals and clustered into general categories by a professional polling company. Contestants attempt to provide an answer which matches these categories and get points according to the proportion of surveyed responses within a matched category.
For example, when we polled 100 people with the same question~(Figure~\ref{fig:intro_fig}), they provided 43 answers involving showering/cleaning, 30 answers mentioning breakfast, and the remainder fell into smaller groups such as locking a door/grabbing keys, saying goodbye, and praying.  In a \ff game, if two participants on a team answered ``grab a shower'' and ``eggs and coffee'', they would receive 73 points for providing answers which matched these two large categories.   We suggest that this is an appealing paradigm for such question answering tasks where a wide range of acceptable answers exist, as it encourages both highly popular answers as well as wide coverage over the range of good answers.  

We frame this task as a generative evaluation task in which a model outputs a ranked list of answers to a given question.  Each answer string is then matched to one or more clusters of reference answers for that question. Matching an answer cluster gives the model a score equal to the cluster size.
Our evaluation metrics (\S~\ref{eval}) reward models which provide the most common answers, while also measuring the model's ability to provide a diverse set of answers in order to match all the answer clusters.
While such an approach can penalize a correct model prediction when it does not match an existing reference answer, we counter this issue by (a) gathering and clustering a large number of reference answers, and (b) utilizing methods of matching non-exact matches, such as WordNet~\citep{miller1995wordnet} and contextual language models such as RoBERTa \citep{liu2019roberta}.  Generative evaluation approaches are also used in other NLP tasks 
such as summarization \cite{radev2003evaluation} and translation \cite{callison-burch-etal-2010-findings}.

We evaluate on a set of competitive baseline models ---  from QA models powered by large masked LMs such as \bert, to the direct prediction of answers in a language-modeling paradigm using a large \gpt LM \cite{radford2018improving}, as well as \gpt fine-tuned upon the training data. While most models perform quite poorly at this challenging task, when \gpt was fine-tuned using the \ff training set its performance did improved drastically, although remaining significantly below the score of human-level performance.



The contributions of this paper are as follows.
\begin{enumerate}
    \item We introduce a large-scale QA dataset of 9.7k questions regarding common sense knowledge of prototypical situations with 7-8 labeled answer categories per question, and a corresponding evaluation set of 15,400 crowd-sourced human judgments over 154 unseen questions.
    \item We present methods for robust evaluation of this task to encourage models to provide diverse answers covering all plausible answer categories.
    \item We evaluate against a range of plausible baselines, showing that while large contextualized language models fine-tuned on this data can perform well at the task, a meaningful gap still exists between model and human performance, suggesting room for improvement. 
\end{enumerate}

\section{Dataset Creation and Analysis} 
\subsection{Training Corpus Collection}
\label{scraping_methods}

A number of fan websites exist which have transcribed \ff questions and answer clusters.  We use publicly available text from two such websites to provide a training dataset on this task.\footnote{Scraping details and site names are provided in the datasheet (following \citet{gebru2018datasheets}) provided with the data}  Well over 10,000 questions (with answer clusters) were collected, and a set of 9,762 questions remained after filtering, quality control, and de-duplication.

That filtering included the omission of questions that were taxonomic in character rather than probing common sense knowledge, such as \textit{name a vegetable}, as well as the omission of questions encoding stereotypes.   A small set of training instances which ascribe specific stereotypes or expectations to a particular group or gender --  such as \textit{``name something little boys love to build models of'}' -- were separated from the main training data set to avoid encouraging trained models to learn such biases~\footnote{Criteria for exclusion are listed in the appendix}. We note, however, that common sense questions may carry a wide range of more nuanced culturally-specific information and biases. Studying the bias in such datasets, and natural stereotypical biases which pre-trained language models have been shown to have~\cite{sheng2019woman}, would be a valuable topic of future work.



\subsection{Test Corpus Collection}
\label{test_set_methods}
In order to establish a rich, open-ended answer generation task, we created new questions similar to those seen in the training set, collected 100 answers for each question\footnote{Each worker, on average, provides 41 judgments, and 5 cents per judgment.} from the crowd-sourcing platform FigureEight\footnote{Now \url{https://appen.com/}.} and manually clustered them. 
Because we gathered large sets of possible answers and clustered them, the evaluation set represents rough distributions over the expected raw string answers for each question, thereby increasing the ability to recognize any way of expressing one of those answers.

We attempted to make sure that this set of new questions maintained the same domain and the same common sense reasoning seen in the training data. In order to maintain similarity to existing questions, these questions were created by removing a set of questions from the scraped data and perturbing important aspects, making sure that the perturbations were sufficient to meaningfully change the answer set (thus being similar to the ``counterfactually augmented'' permutations of \citet{kaushik2019learning}).  For example, given an existing question of ``\textit{Name something a person might forget to put on if they leave the house in a hurry.}'', changes of polarity and events would derive a related question ``\textit{Name something that people usually do before they leave the house for work}''.  Deriving such unseen test questions was especially important to avoid the risk of having a publicly-available question be included in the training data for contextual language models; by making new data, we can be confident that any high-performing model has not yet seen the data.  In order to control the quality of perturbed questions, the quality of each each perturbed question was scored by four experts (criteria listed in the appendix), and only the top-scoring questions were used to build the evaluation set.

We then created tasks on FigureEight for each selected question to be answered by 100 workers. To match the training data (which is inherently grounded in US culture), we limited workers to US locations. Low-quality workers were automatically detected through test questions during annotation, and the clustering pass provided a second manual quality control check. This left us with 154 questions which we split into a test set and development set of 102 and 52 respectively.



\subsection{Answer Clustering}
\label{clustering}

Each list of 100 raw string answers was manually clustered by two different experts familiar with the task. Clusters were assigned separately and then compared, and a final clustering was agreed on.\footnote{The four total expert annotators annotated a random set of 10 questions together to calibrate their clustering granularity. Furthermore, two annotator’s results are aggregated by a third person to reduce bias.}  During this clustering phase answers could be marked as invalid as well  --- most commonly, either due to low-quality annotations or a clear misunderstanding of a question.   In order to keep these clusters roughly similar to the granularity of answers used in the training data and to avoid low-quality evaluation we eliminated questions for which the 8 most popular clusters did not contain at least 85 of the 100 responses.



Since each set of answers was clustered twice and adjudicated, we measure the agreement with a cluster agreement metric BLANC~\cite{recasens2011blanc,luo2014extension}, an extension of the Rand index used to score coreference clustering. Using this, the similarity between the clusters produced by any two annotators averaged out to a BLANC score of 83.17, suggesting a coherent amount of agreement regarding the clustering of answers.



\subsection{Analysis of the Dataset}
\label{data_exploration}

\begin{table*}
\begin{tabular}{l  l  c }\\\toprule
Question & Example Answers & Types \\\toprule 
Name a profession where you might be fired if you lost your voice&  radio host , teacher &  3, 4, 6 \\ 
Name something a boy scout might learn.  & knot tying, camping & 	2, 5, 6 \\
Name a bad sport for someone who is afraid of the water. & diving, water polo & 	1, 3 ,6 \\
Name something a monk probably would not own. & weapons, smartphone
& 2, 4, 6 \\ 
Name something parents tell their kids not to do & steal, smoke & 1, 2, 4, 6 \\
Name a reason why someone would wear gloves & cold weather, cleaning & 2, 3 \\\bottomrule
\end{tabular}
\caption{\label{question_examples_for_analysis} Examples of questions from collected (top 3) and crowd-sourced (bottom 3) development sets, characterized with reasoning types described in \S~\ref{data_exploration}}
\end{table*}

The data presented here involves a range of different types of common sense knowledge.  To explore the distribution of different kinds of reasoning, and to test whether that distribution of reasoning varied between the publicly available data and the crowdsourced development and test set, 
we propose a small inventory of six types of common sense reasoning.  

We are not aware of an agreed-upon typology of all commonsense reasoning types. Categorizations of different types of commonsense reasoning exist~\citep{lobue2011types,boratko2018systematic}, but since each provided categorizations needed for specific tasks (RTE and the ARC dataset, respectively), neither fully covered the range of commonsense types seen in the current work. After consulting both those prior works and a separate part of the training data, we characterize the data into the following six types.

These types consist of (1) \textsc{Mental or Social Reasoning}, (2) \textsc{Knowledge of Prototypical Situations} which one is familiar with,  (3) \textsc{Reasoning about novel, complex events}, (4) \textsc{Negation and exceptions} and understanding their consequences, (5) \textsc{Specific Entity knowledge} of named people, locations, or organizations, and finally (6) \textsc{Knowledge of habitual activities} of specific occupations or types of entities. 


Following other characterizations of reasoning type~\citep{lobue2011types,boratko2018systematic}, we annotated a random sample of questions (25 from dev and 25 from train) using six basic common sense reasoning categories in order to provide a simple approximation of the distribution over reasoning types contained in the data.
Table \ref{question_examples_for_analysis} illustrates examples of questions with these types, and one can see the frequency of each type used in Table \ref{reasoningdistribution}.  The counts shown for each dataset illustrate that while the creation methodology varied between the two resources, the kind of common sense reasoning tasks evaluated by these models is quite similar between the two corpus types.  The greatest difference to note is that the crowd-sourced data makes less use of questions regarding specific entities, which were avoided as they tended to involve fact-based world-knowledge rather than common sense reasoning. 

\begin{table}[!h]
\small
\begin{tabular}{l c  c}\\\toprule
Reasoning type & Scraped Dev & Crowd-sourced \\ \toprule
Mental/Social  & 16\% & 12\% \\
Prototypical Events  & 68\% & 80\% \\ 
Event Reasoning & 28\% & 40\% \\
Negation  & 12\% & 20\% \\
Specific Entities  & 20\% & 4\% \\ 
Habitual Activity & 40\% & 24\% \\\bottomrule

\end{tabular}
\caption{\label{reasoningdistribution} Percentage of questions utilizing each reasoning type}
\end{table}

\section{Evaluation}
\label{eval}


\begin{figure*}[h!]
\includegraphics[width=\textwidth]{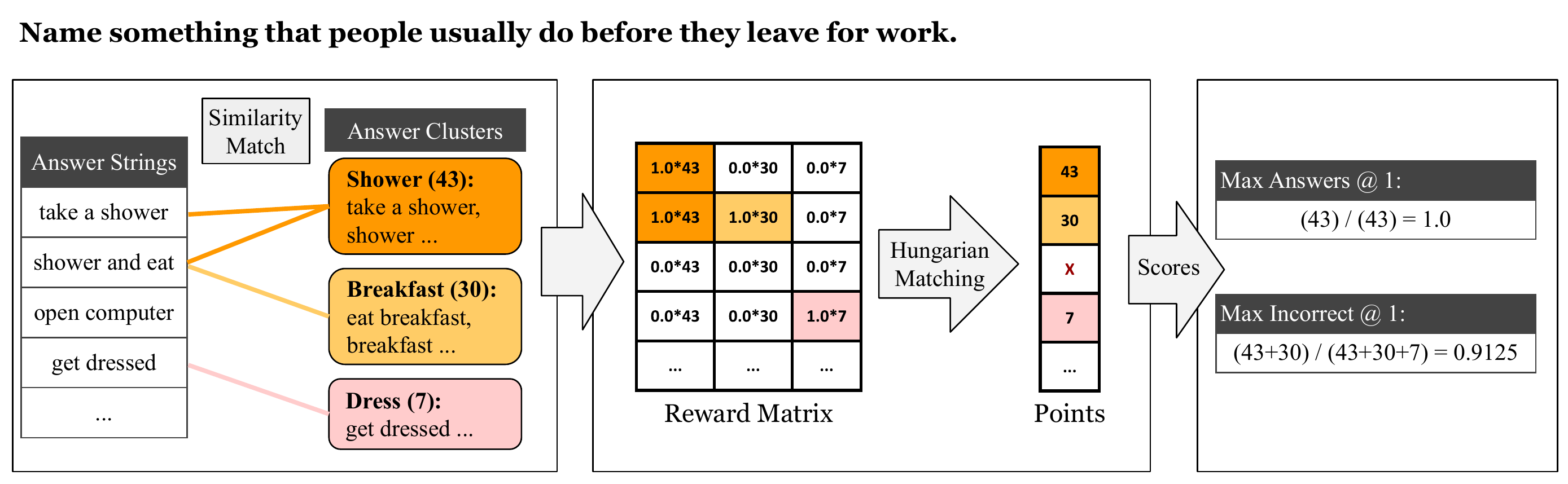}
\caption{Example steps for evaluating a ranked list of answers}
\label{fig:evaluation function}
\end{figure*}

We present a number of methods for evaluating system-generated answers against these sets of clustered answers.  In each, models are evaluated by providing a ranked list of answers in response to a question.  These answers are then compared to the set of reference answers for that question and scored based upon how similar they are to the known answers.  While one might instead convert question-answer pairs into a multiple-choice paradigm by generating negatives, it is difficult to generate good negative examples, and the quality of a dataset can be compromised if such examples are either too easy or easily identified using biases in the negative example generation process ~\citep{mostafazadeh2016corpus,zellers2018swagaf,talmor2018commonsenseqa,schwartz2017effect,gururangan2018annotation,poliak2018hypothesis}.

We outline here our proposed method for scoring these ranked lists of predicted answers. The dataset ground truth is a ranked list of clusters of answers, including weights(cluster sizes) associated with each cluster. A first component in such an evaluation is to match each answer to an existing cluster of answers, if any cluster is acceptable. We try both simple methods such as exact match as well as more flexible ways of matching to clusters, such as using synonyms from WordNet \cite{miller1995wordnet} or a vector-based similarity method using RoBERTa~\citep{liu2019roberta}.  The second component in this generative evaluation is to provide an overall score for the entire ranked list of answers by mapping individual answers to answer clusters or marking them wrong.  
Scoring answers against clusters alone does not take into account the ranking. To that end, we propose two different metrics, one similar to hits@k in traditional information retrieval task and one which limits the number of incorrect answers, which is closer to how humans are typically evaluated on this task.

In each case the score reported is calculated as a percentage of the oracle score.
Both proposed methods of scoring reward models which provide a diverse set of guesses to a given query and penalize models which provide many variations of the same answer. (See figure \ref{fig:evaluation function} for a general idea of the steps involved.)




\subsection{Matching Answers to Clusters}
\subsubsection{Exact Match}
\label{sub:exact}

In our simplest way of matching answers to clusters, we compare each answer with the answer strings from crowd-source workers for a given cluster, returning a score of $1$ if it matched any string in the cluster and returning $0$ if not. By construction, therefore, a given answer string will match at most a single cluster with this method.

\subsubsection{WordNet Similarity}
\label{sub:wn}
Reasonable answer strings may be incorrectly marked as wrong with an exact string match, even when they are clear synonyms of a reference answer. \textsc{METEOR}~\citep{banerjee2005meteor, lavie2009meteor} addressed similar issues in machine translation via stemming and synonym matching. We take a similar approach, tokenizing a proposed answer string and comparing it to the tokenization of the answers in each answer cluster. Since some words in WordNet are multi-word phrases (eg. ``chewing gum") we furthermore perform this matching on all possible partitions of the tokenization. For each answer in an answer cluster we return the maximum (over all possible partitions) of the average number of  matched tokens. The assignment of answers to clusters proceeds as in the exact match case. Further details are included in the appendix.

\subsubsection{RoBERTa Similarity}
\label{sub:clm}
Recent works in MT evaluation~\citep{zhang2019bertscore,sellam2020bleurt} used pre-trained language models to compare predictions to reference answers.  We implement a simple version of such vector-based comparisons, but this current task differs in that we assign each predicted answer to a particular cluster of correct answers, or decide whether to reject the answer.  As clusters vary in size and specificity we cannot determine a universal threshold for how similar a mention must be to a cluster. Instead, we train a small classifier in L2 distance space for each answer cluster in order to decide membership in that answer cluster.  We do this by obtaining a vector representation of each answer from RoBERTa~\citep{liu2019roberta}, concatenating each answer with the question, and taking the mean of answer token representations.  For each cluster we train a small one-vs-all classifier over the 100 answers to that question, predicting membership in that cluster (using gaussian process regression~\citep{williams1996gaussian} with an RBF kernel).  At test time, a given answer is assigned to the highest-scoring cluster, as long as its likelihood of membership exceeds a minimum probability threshold, set at 0.1.  Such an approach allows us to match answers to clusters while omitting answers which do not match existing clusters.  
\subsection{Evaluating Diverse Lists of Answers}
\label{sub:diversity}
As mentioned previously, we want to design evaluation metrics that favor models which take into account the ranking while still covering all plausible answer categories. We first compute an alignment score between each answer in the ranked list and each of our answer clusters. After computing the alignment scores between all pairs of answers and clusters we create a reward matrix where, for each answer and cluster, we assign a reward equal to the cluster size if the alignment score was a $1$ and $0$ otherwise. We employ the Hungarian matching algorithm \cite{kuhn1955hungarian,munkres1957algorithms} to compute the exact optimal matching of answers to clusters based on this reward matrix, so that an answer is assigned to only one cluster. It is worth noting that a model which produces a ranked list of answers only in one cluster will do worse than a model which maximally covers all plausible clusters. Lastly, to make the comparison between lists of different lengths uniform, we propose the following metrics.
\begin{enumerate}
    \item \textsc{Max Answers@$k$} limits the total number of answers allowed to up to $k$ answers.\footnote{Note that since our scores are always calculated as a percentage of the max score one could receive, \textsc{Max Answers} is slightly different than hits@k in this setting.}
    \item \textsc{Max Incorrect@$k$} allows unlimited answers, but stops after $k$ unmatched answers are provided.
\end{enumerate}

In both conditions, we report the score as the percentage of the max score one could receive given that number of guesses, and only give credit for a given cluster once.

\section{Baselines}
\label{baselines}
\begin{table*}[]
\small
\centering
\begin{tabular}{c|c|c|c|c|c|c}
\toprule
\multicolumn{3}{c|}{\textbf{Metrics \%}} &  \textbf{QA Model} & \textbf{\gpt} & \multicolumn{1}{c|}{\textbf{\begin{tabular}[c]{@{}c@{}}\gpt\\  Fine Tune\end{tabular}}} & \textbf{\begin{tabular}[c]{@{}c@{}}Human\end{tabular}} \\ \midrule
\multirow{8}{*}{\textbf{\begin{tabular}[c]{@{}c@{}}Exact\\ Match\end{tabular}}} &  \multirow{4}{*}{\textbf{Max Answers}} 
 & 1 &  2.1 & 5.6 & 29.4 & \textbf{78.4}     \\
 &  & 3   & 4.4 & 15.9 & 37.6 & \textbf{74.4}  \\
 &  & 5 & 6.8 & 18.3 & 40.1 & \textbf{72.5}  \\
 &  & 10 &  11.0 & 23.2 & 45.9 & \textbf{73.3} \\ \cmidrule{2-7} 
 & \multicolumn{1}{l|}{\multirow{3}{*}{\textbf{Max Incorrect}}} 
& 1 &   0.8 & 3.3 & 18.7  & \textbf{55.8}  \\
 & \multicolumn{1}{l|}{} & 3  & 3.6 & 15.1 & 35.0 & \textbf{69.4}  \\
 & \multicolumn{1}{l|}{} & 5 & 6.4 & 19.3 & 41.1 & \textbf{72.4}   \\ \midrule
 
 \multirow{8}{*}{\textbf{\begin{tabular}[c]{@{}c@{}}WordNet\\ Similarity\end{tabular}}} &\multirow{4}{*}{\textbf{Max Answers}} 
 & 1  & 3.4 & 6.2 & 36.4 & \textbf{78.4}  \\
 &  & 3  &  6.4 & 18.5  & 44.4 & \textbf{76.8}  \\
 &  & 5  &  9.1 & 23.0 & 46.6 & \textbf{76.0}   \\
 &  & 10 &  15.7 & 30.5 & 53.5 & \textbf{77.0}   \\ \cmidrule{2-7} 
 & \multicolumn{1}{l|}{\multirow{3}{*}{\textbf{Max Incorrect}}} 
& 1 &  1.4 & 4.3 & 26.1 & \textbf{59.0} \\
 & \multicolumn{1}{l|}{} & 3 & 5.3 & 17.9 & 41.7 & \textbf{74.0} \\
 & \multicolumn{1}{l|}{} & 5 & 8.4 & 24.2 & 48.2 & \textbf{77.9}  \\ 
 
 \midrule
 
 \multirow{8}{*}{\textbf{\begin{tabular}[c]{@{}c@{}}RoBERTa\\ Similarity\end{tabular}}}  
 & \multirow{4}{*}{\textbf{Max Answers}} 
 & 1 & 49.1 & 38.7 & 55.0 & \textbf{81.2}  \\
 &  & 3  & 53.3 & 48.8  & 60.7 & \textbf{78.9}  \\
 &  & 5  &57.1 & 52.0 & 63.0 & \textbf{80.1}   \\
 &  & 10 &  65.0 & 60.5 & 71.2 & \textbf{83.5}   \\ \cmidrule{2-7} 
 & \multicolumn{1}{l|}{\multirow{3}{*}{\textbf{Max Incorrect}}} 
& 1 & 49.1 & 38.7 & 55.0 & \textbf{81.2} \\
 & \multicolumn{1}{l|}{} & 3 & 53.3 & 48.8 & 60.7 & \textbf{78.9} \\
 & \multicolumn{1}{l|}{} & 5 & 57.1  &52.0 & 63.0 & \textbf{80.1}  \\ 
 \bottomrule
\end{tabular}
 \caption{Results on the annotated test set. Scores are normalized by the maximum score obtainable with that number of guesses, and therefore may go down as k increases}
\label{table:test-results}
\end{table*}
We explore three baseline models for this task: a QA-based model which retrieves related posts in a discussion forum for each question, a language-modeling baseline which examines how well modern pre-trained language models do at directly producing the answers, and a fine-tuned version of the language-model baseline.

\subsection{Question-Answering Baseline}

As this dataset is in the form of questions and answers it may be treated as a QA dataset, although the content is far from the fact-based data usually modeled in QA tasks.  As the training set only shows answers out of context, one must use distant supervision in order to train a QA model on the data, a well-explored situation in modern QA work ~\cite{joshi2017triviaqa}.  
Unlike factoid-based QA, one may expect a limit in the performance of such QA models for common sense reasoning, as common sense data is well-known to have a \textit{reporting bias}~\cite{gordon2013reporting} wherein many facts that are part of the common ground of known knowledge are less likely to be stated. 


To train a model in this approach, we collected up to 20 documents for each of the 9.7k questions in the \ff training dataset by using a web search for each question constrained to Reddit.  This resulted in a set of 85,781 Reddit posts total.   Searches were constrained to Reddit in order to focus upon advice and personal narratives which might discuss common sense questions.  For any post matching that query, any strings matching an answer to that question in the training data would be treated as a positive example for the QA model. The QA model used was the ``Bert for QA'' implementation within the Hugging Face Transformers package~\cite{Wolf2019HuggingFacesTS}; training details, and examples of the kind of noisy training data generated through this process, are provided in the appendix.  

At test time documents were obtained by searching for the question in a google search restricted to Reddit, and the QA model was run on that set, taking the 20 best answers in context as possible answer strings.  Those best answer strings from each passage were combined together, summing scores for identical strings, to provide a ranked list.

\subsection{Language Model Baseline}
We also report a language model generation baseline, due to the improved representation power of modern language models and recent evidence of their power in modeling common sense reasoning tasks~\citep{adamtacit2020,tamborrino2020pre}. The baseline is performed using the AI2 \gpt large model~\cite{radford2019language}  (specifically, the Hugging Face PyTorch implementation~\cite{Wolf2019HuggingFacesTS}). We perform both a zero-shot evaluation and an evaluation after fine-tuning with using our training data.

Because the original \ff prompts are not structured as completion tasks, we transform the original question by hand-designed transformation rules in order for it to be compatible with the \gpt training data. E.g ``Name something people do when they wake up.'' $\rightarrow$ ``One thing people do when they wake up is ...''. The hand-designed rules are including in the appendix. The transformed questions are used as input to the language model, and \gpt finishes the sentence. The reported fine-tuning result is trained on the scraped training corpus and the best model selected based on performance on our annotated development set. Training details and parameter setting for the model is provided in the appendix.

In order to generate diverse answers for a given sentence we use Nucleus Sampling ~\cite{holtzman2019curious} as our decoding method. We get 300 sampled answers for each question and group them by counts, returning a ranked list of 20 answers from most to least common.

\subsection{Human Performance}

To measure human performance against such models, we collected 30 additional human responses per question with the same setup in collecting test data and aggregated them by counts, just as the sampled answers from \gpt models were ranked.  
The last column in table~\ref{table:test-results} reports this human performance. We can see that the best-performing automatic system is still meaningfully behind human performance in all metrics.

\section{Discussion and Analysis}

Table~\ref{table:test-results} shows the results of the baseline models using different measures of similarity, and different measures for the \textsc{Max Answers} and \textsc{Max Incorrect} metrics.  One can see that \gpt without fine-tuning outperforms the baseline QA implementation, and fine-tuned \gpt outperforms both, but a large gap still remains between human performance and any of the baselines, even on the generous RoBERTa-based similarity metric.
The human baseline scores are relatively stable regardless of which similarity metric is used, whereas the model scores change drastically (most significantly for the QA model) as more generous similarity metrics are used.  We suggest that WordNet Similarity be used as the primary similarity metric as it strikes a reasonable balance between precision and recall, as discussed in \S~\ref{sec:Score Function Comparison}.


\subsection{Knowledge Base Comparison}

To show the dataset indeed containing meaningful commonsense knowledge, we did an additional analysis between our dataset and ConceptNet. ConceptNet~\cite{speer2017conceptnet} is a knowledge base containing triples related to common sense which has been shown to be helpful for various downstream tasks~\citep{zhong2019improving, wang2019improving} and conversational text generation~\citep{diverse2020,conversation2020}. We evaluate its potential relevance to this task by evaluating how often a (question, answer cluster) pair has a possible matching triple within ConceptNet. We extract a list of keywords from the question and a ground-truth answer string (by removing stop words) and similarly extract keywords from the head and tail of each ConceptNet relation.  	We then evaluate whether a given question-answer pair has potential ``coverage'' in ConceptNet by checking whether a keyword in the question is related to a keyword in the answer.
For example, given the question ``\textit{Besides music, name something you might hear on a morning radio show}'' and the answer ``\textit{weather report}'', we would find the triples \textit{(listen to radio, Cause, you hear local weather report)} and \textit{(listen to radio, HasSubevent, hear weather report)}.    
\begin{table}[tb]
\centering
\begin{tabular}{@{}l|ccc@{}}
\toprule
 & \textbf{Precision} & \textbf{Recall} & \textbf{F1} \\ \midrule
\textbf{\begin{tabular}[c]{@{}l@{}}Exact \\ Match\end{tabular}} & \textbf{1.0} & 0.466 & 0.636 \\ \midrule
\textbf{\begin{tabular}[c]{@{}l@{}}WordNet\\ Similarity\end{tabular}} & 0.996 & 0.581 & \textbf{0.734} \\ \midrule
\textbf{\begin{tabular}[c]{@{}l@{}}RoBERTa\\ Similarity\end{tabular}} & 0.762 & \textbf{0.661} & 0.708 \\ \bottomrule
\end{tabular}
\caption{\label{metric_measure} Measurement of different score function against human cluster assignment. }
\end{table}
By this measure, we find that 24.3\% of the answer clusters in our development set have some match within ConceptNet.  
This suggests that a common sense KB might provide a useful resource for this task, however ConceptNet has a large number of relations with no direct ability to provide a ranking and thus we exclude such a model from our baseline comparisons.
A similar analysis shows that the human baseline match 46.5\% of the clusters, whereas a list of 20 top answers from the fine-tuned \gpt model match 30.3\%.

\begin{table*}[htb]
    \centering
    \begin{tabular}{@{}l|llll@{}}
        
        

    \toprule
    \textbf{Prompt} & \multicolumn{4}{l}{\textit{\textbf{Name something around the house that's often replaced.}}} \\ \midrule
    Human & \colorbox{largest}{light bulbs} & \colorbox{second}{toilet paper} & \colorbox{small}{furniture} & \colorbox{small}{food} \\
    \gpt & TV & refrigerator & fridge & \colorbox{third}{trash} \\
    \begin{tabular}[c]{@{}l@{}}\gpt Fine Tune\end{tabular} & dishes & \colorbox{second}{toilet} & kitchen & \colorbox{small}{furniture} \\
    QA & tune & time & name & song \\ \midrule \midrule
    
    \iftrue 
    \textbf{Prompt} & \multicolumn{4}{l}{\textit{\textbf{Name something a monk probably would not own}}} \\ \midrule
    Human & \colorbox{largest}{gun} & \colorbox{small}{wife} & \colorbox{largest}{knife} & \colorbox{second}{pornography} \\
    \gpt & \colorbox{largest}{gun} &    \colorbox{third}{car} & \colorbox{largest}{sword} & \colorbox{third}{motorcycle} \\
    \begin{tabular}[c]{@{}l@{}}\gpt Fine Tune\end{tabular} & \colorbox{largest}{weapon} & \colorbox{largest}{sword} & \colorbox{third}{car} & \colorbox{third}{cell phone} \\
    QA & arch & everything & togashi & power \\ \midrule \midrule
    
    \else 
    \textbf{Prompt} & \multicolumn{4}{l}{\textit{\textbf{Name a complaint people have about their gym instructors}}} \\ \midrule
    Human & none & \colorbox{largest}{pain} & instructors & damaged machines \\
    \gpt & they're nice & \colorbox{largest}{strict} & nice & listen \\
    \begin{tabular}[c]{@{}l@{}}\gpt Fine Tune\end{tabular} & \colorbox{second}{they're slow} & they're friendly & \colorbox{second}{slow} & \colorbox{third}{they're rude} \\
    QA & \colorbox{largest}{hard} & everyone & coaches & going \\ \midrule
    \fi
    \multicolumn{5}{l}{ \colorbox{largest}{largest cluster}  \quad \colorbox{second}{cluster 2} \quad \colorbox{third}{cluster 3} \quad \colorbox{small}{smaller clusters}} \\ \bottomrule

    \end{tabular}
        \caption{Top responses from human and model predictions for each prompt, color-coded with the answer cluster they might be aligned to}
    \label{tab:modeloutputs2}
\end{table*}   
\subsection{Score Function Comparison}
\label{sec:Score Function Comparison}
In order to compare the various similarity functions outlined in \S~\ref{eval}, we manually annotated answers -- from both the human baseline and fine-tuned \gpt outputs -- to the correct answer clusters. Four annotators separately mapped each answer string to an existing cluster.


Table \ref{metric_measure} measures how well different similarity functions performed in comparison to the manual human cluster assignment.  Precision in this context measures how often an answer assigned by the automatic similarity measure is correctly assigned; recall measures how often an answer which a should be assigned to a cluster is correctly assigned. Unsurprisingly, exact match has perfect precision in this context, but has relatively low recall.  WordNet similarity increases recall while adding very little false positives.  As was hoped, RoBERTa similarity does dramatically increase how often an answer is mapped to the correct cluster, but does so at the expense of a large loss in precision; we therefore suggest that the WordNet similarity is the safest evaluation option.



\subsection{Error Analysis}


To provide some notion for the tendencies of different models on this task we provide actual model outputs in Table \ref{tab:modeloutputs2}.
One can see that, before fine-tuning, \gpt results are often acceptable and plausible situations (e.g. refrigerators might be replaced), but can fail to answer the specific criteria requested by the prompt. In contrast, the QA-based model is much noisier -- occasionally providing very good answers, but often (as in the examples  provided) failing to find answers that are even plausible.  Fine-tuned \gpt, in contrast to both, clearly learns to actually focus upon the expected format and details of such prototypical activities, however it fails in situations where a high-scoring answer would be very rarely discussed, such as knowing that light bulbs are commonly changed around the house. 

\section{Related Work}
\label{relatedwork}

A wide variety of common sense reasoning datasets address related topics. Many datasets cover physical and spatial reasoning~\citep{bisk2019piqa}, social common sense~\citep{sap2019socialiqa}, and common sense understanding of plausible sequences of events~\cite{zellers2018swagaf,zellers2019hellaswag,huang2019cosmos,bhagavatula2019abductive,sap2019atomic} or understanding of the entailments of a sentence~\cite{zhang2017ordinal,bowman2015large,roemmele2011choice,levesque2012winograd}. There is also a long history of work in modeling scripts and frames~\citep{schank1977scripts,chambers2009unsupervised,fillmore1976frame,ferraro2016unified,weber2020causal}, which is related to the current focus on prototypical situations.  

Recent works have also sought to characterize the ability of pre-trained language models to understand common sense reasoning, showing such models perform well at common sense reasoning tasks even without fine-tuning, allowing one to explore the common sense reasoning inherent in those models~\citep{tamborrino2020pre,adamtacit2020}. Of particular relevance to the current work, \citet{adamtacit2020} explored the ability of pre-trained models to predict \textit{stereotypic tacit assumptions}, generalizing about entire classes of entities with statements such as ``everyone knows that a bear has $\rule{1.5cm}{0.15mm}$''.

Interestingly, ProtoQA is not the first time \ff has been referenced in the commonsense literature.  Common Consensus \citep{Lieberman07commonconsensus:} was a web-based game created with the intention of being a self-sustaining platform to collect and validate commonsense knowledge based on human goals.  Prior work had established the idea of using online games to simultaneously entertain and collect commonsense knowledge \citep{Ahn06verbosity:a}, however the authors of Common Consensus found that the format of \ff questions was more amenable to high-quality commonsense knowledge acquisition.
Common Consensus serves as an excellent proof of concept for future gamification of the style of data presented in this dataset.

ProtoQA differs from other datasets in three different ways:
\begin{enumerate}
    \item ProtoQA focuses on proto-typical situations. Humans can agree about the details and characteristics of a prototypical event or situation due to commonalities in their shared lived experiences, cultural norms and expectations. This rough agreement extends beyond an agreement on a single top response and that’s why our task and evaluation values diversity of answers.
    \item The evaluation ProtoQA is a generative evaluation task where a model has to output a ranked list of answers, ideally covering all prototypical answers for a question.
    \item ProtoQA has a large number of annotations for each example which makes the generation evaluation possible.
\end{enumerate}

\section{Conclusion}
\label{conclusion}
We have presented a new common sense dataset with many novel features.  The collection of a large set of raw answer strings and further clustering of these strings facilitates a generative evaluation method, enabling actual use of trained models to answer real common sense questions.  The inclusion of counts over clusters of answers provides a very rich dataset for training and evaluation. As shown in table \ref{table:test-results}, existing fine-tuned state-of-the-art models have a way to go before modeling the distribution of this common sense data. 

In addition to the elements of this task which are appealing from a common sense modeling perspective, the inherent appeal of this task to humans opens a number of possibilities for future data collection and evaluation.  Millions of people have played phone-based games based upon this same premise\footnote{Based on downloads of \url{https://play.google.com/store/apps/details?id=com.umi.feudlive}}, and prior works have obtained valuable annotations from trivia game participants~\citep{rodriguez2019quizbowl}.  This dataset lays the foundation for larger-scale data collection which leverages people's natural interest to encourage high-quality answers to more common sense questions.


\section*{Acknowledgments}

We thank the IESL and NLP lab at UMass Amherst for their efforts in assisting with data collection. This work was supported in part by the Center for Intelligent Information Retrieval and the Center for Data Science, in part by the Chan Zuckerberg Initiative under the project Scientific Knowledge Base Construction, and in part by DARPA. Any opinions, findings and conclusions or recommendations expressed in this material are those of the authors and do not necessarily reflect those of the sponsor.

\newpage
\bibliographystyle{acl_natbib}
\bibliography{anthology,emnlp2020}

\clearpage

\newpage

\appendix
\section{WordNet Similarity Function}
\label{sec:WordNet Similarity Function}
\begin{enumerate}
\item  Let $S$ be the set of synsets in WordNet, and let $S(x)$ be the set of synsets associated with the string $x$.
\item Let $\operatorname{SynsetSim}(X,Y):S\times S \to [0,1]$ be a score for synset similarity, eg.
\[
\operatorname{SynsetSim}(X,Y) := \begin{cases}
1 \quad &\text{ if } X = Y,\\
0 \quad &\text{otherwise}.
\end{cases}
\]
\item A given string may corresponse to multiple synsets. Given two strings $x$ and $y$ we define
\begin{multline*}
\operatorname{SynsetsScore}(x,y) = \\\max\{\operatorname{SynsetSim}(S_x, S_y) : S_x \in S(x), S_y \in S(y)\}.
\end{multline*}
\item Some valid answer strings may not correspond to a synset at all, so we define
\begin{multline*}
\operatorname{SubstringScore} = \\
\max(\operatorname{SynsetsScore}(x,y), \operatorname{ExactMatch}(x,y))
\end{multline*}
\item Some answers are several words long, and therefore won't map to a synset even if some substring would. To account for this, we tokenize and strip stopwords from both the predicted and ground-truth answer strings. To compare these sets of tokens $A,B$ we let $M(A,B)$ be the set of all possible (partial) matchings between elements in $A$ and $B$, and then define
\begin{multline*}
\operatorname{TokensScore}(A,B)\\= \max_{m\in M(A,B)} \frac{\sum_{(a,b) \in m} \operatorname{SubstringScore}(a,b)}{\max(|A|,|B|)}
\end{multline*}
\item We repeat this process for every element in an answer cluster $C$, which is a set of strings obtained from the survey, and then set the overall score for this answer cluster to be
\begin{multline*}
    \operatorname{WordNetScore}(x,C) = \\
    \max\{\operatorname{TokensScore}(T(x), T(y)) : y \in C\}
\end{multline*}
\end{enumerate}

\begin{remark}
Fully tokenizing the input has the potential to lose information, since some WordNet clusters are labeled with multiple words. Consider comparing ``chewing gum" with ``gum". The above process would assign this a score of $0.5$, because tokenizing yields [``chewing", ``gum"], however ``chewing gum" is, itself, in the same WordNet synset as ``gum''. The solution to this problem in general is to compare all possible \emph{partitions} of the tokens, and define the overall $\operatorname{PartitionsScore}$ to be the maximum among all pairs of possible partitions for the predicted answer and the ground-truth string. We replace the $\operatorname{TokensScore}$ with this $\operatorname{PartitionsScore}$ to capture such situations.
\end{remark}

With a scoring method as described, it is possible for an answer to receive a positive score for multiple clusters. We take the following approach:
\begin{enumerate}
    \item Round the scores to \(\{0,1\}\) to make a "hard" cluster decision.
    \item For a given question, if some predicted answers match with multiple clusters, we choose the maximum matching with respect to the final score. 
\end{enumerate}




\section{\gpt Transformation rules}
\begin{table}[h]
\begin{tabular}{|l|l|}
\hline
Original Sentence & Transformed Sentence \\ \hline
Name something ... & One thing ... is \\
Tell me something ... & One thing ... is \\
Name a/an ... & One ... is \\
How can you tell ... & One way to tell ... is \\
Give me a/an ... & One ... is \\ \hline
\end{tabular}
 \caption{\label{transformation}Transformation rules from original question sentence to \gpt format sentence}
\end{table}
In order to make the question more natural for \gpt model to answer, we use rule in Table \ref{transformation} to re-write the questions.

\section{Criteria for test question acceptance}

When creating new questions using the perturbation method described in \S~2.2, we scored each question with the following criteria in mind:
\begin{itemize}
\item Most people are expected to be able to answer. 
\item The answer set category is relatively small; less than eight big categories of different answers. 
\item The question is hard for systems relying on co-occurrence patterns to answer, e.g., \bert 
\item The answers to the question are not too culturally dependent (e.g., we want to avoid questions such as \textit{Name a dish made with ground meat}).
\item Not accidentally re-creating a well-explored question:  We then searched all Family Feud data to ensure that no questions were being re-created, and searched online to make sure no obvious lists of answers can be found via search with Google.  E.g., if we create a question and the top search for that question is a list of answers to that question, regardless of origin, we remove the question. 
\end{itemize}

\section{Criteria for stereotypical bias issues}

We define a relatively strict measure for stereotypical bias, primarily to avoid having overly problematic examples; we expect that more nuanced issues of stereotypes are common in the data, but are not as easy to measure with an all-or-nothing measure. We rule out questions if they match any of the following:
\begin{enumerate}
\item Attaining the right answer requires stereotypes regarding what activities are affiliated with each gender (e.g., that only girls play with dolls)
\item Questions that measure activities a particular gender would be proud or embarrassed to do.
\item We could not find any questions addressing race, sexual orientation, religion, or national origin, but these were searched for and would have also been removed if found. 
\end{enumerate}
Types of potentially biased questions which we could not consistently remove from all the training data, but which we note to be worthy of consideration, are:
\begin{enumerate}
\item Questions with heteronormative assumptions (questions about what women like, romantically, in men or vice versa)
\item Questions that can be specific to Western US culture: a vast array of questions would have different distributions over answers if asked to people of specific cultures, where stereotypical foods, greetings, habits, or objects may be different.
\item Questions that reference gender, but which might have similar answer clusters if the gender was removed -- e.g., \textit{Name something your parents always want to know about the man you're dating}.	
\end{enumerate}

\section{QA model details}

For the baseline results reported, we fine-tune the ``Bert for QA'' model of the Huggingface transformers package, v2.6.0 ~\cite{Wolf2019HuggingFacesTS}, using BERT-large-uncased ~\cite{devlin2018bert}.

Table \ref{qamodel} illustrates examples of answer strings for the query ``name something you do at a concert'', illustrating both that such a method finds passages that are relevant to the questions, but also illustrating the kind of noise being introduced by such a distance learning approach.  

\begin{table}[h]
\begin{tabular}{p{6.9cm}}
Q: Name something you do at a concert: \\ 
A: \textit{But you are always expected to \textbf{\textcolor{red}{clap}} for the spalla .} \\ 
A: \textit{I'll often buy a \textbf{\textcolor{red}{drink}} for something to do, or check my email on my phone, or whatever, to kill time . once the band starts i 'm focused on that}  \\
\end{tabular}
\caption{\label{qamodel}Examples of distant-learning positive examples used for training QA baseline}
\end{table}

\section{\gpt model details}

For the baseline results reported, we fine-tune \gpt Large model using the scrapped training data. The parameter for the best performing model is as follows: batch\_size:1, training epoch: 1, gradient accumulation step: 8. The other parameters are the default value in the hugging face implementation. In generation phrase, the temperature is 0.69, top\_p is 0.9, and other parameter values are using the default values. All parameters are tuned using dev data, and searched via greed search. The code will be publicly available upon publication.

\section{Alternative Human Performance Answers}
\begin{table*}[htb]
    \centering

    \begin{tabular}{p{2.9cm} |  p{2.9cm}  p{2.6cm}  p{2.4cm} p{2.4cm} } \hline
    Prompt & \multicolumn{4}{|l}{\textit{Name something around the house that's often replaced.}}\\ \hline
     Single-human ranking  & \colorbox{small}{food} & \colorbox{second}{toilet paper} & \colorbox{small}{paper towels} & \colorbox{third}{garbage bags} \\ \midrule 
     
     \iftrue 
     Prompt & \multicolumn{4}{|l}{\textit{Name something a monk probably would not own.}}\\  \hline
     Single-human ranking  & \colorbox{third}{a fancy car} & {a fancy house} & {too much food} & \colorbox{small}{a bank account} \\ \hline
     
     \else 
     Prompt & \multicolumn{4}{|l}{\textit{Name a complaint people have about their gym instructors.}}\\  \hline
     Single-human ranking  & \colorbox{largest}{Too strict} & \colorbox{second}{Non-puntual} & \colorbox{third}{impolite} & \colorbox{small}{Talkative} \\ \hline
     \fi
     
 \multicolumn{5}{l}{ \colorbox{largest}{largest cluster}  \quad \colorbox{second}{cluster 2} \quad \colorbox{third}{cluster 3} \quad \colorbox{small}{smaller clusters}}  \\ 
        \bottomrule

    \end{tabular}
        \caption{Top three responses from human ranking evaluation for the same data}
    \label{tab:hrank}
\end{table*}  

\begin{table}[h]
\small
\centering
\begin{tabular}{c|c|c|c}
\toprule
\multicolumn{3}{c|}{\textbf{Metrics \%}} &   \textbf{\begin{tabular}[c]{@{}c@{}}Single-Human \\ Ranking\end{tabular}} \\ \midrule
\multirow{8}{*}{\textbf{\begin{tabular}[c]{@{}c@{}}Exact\\ Match\end{tabular}}} &  \multirow{4}{*}{\textbf{Max Answers}} 
 & 1 &  40.5     \\
 &  & 3    & 39.4  \\
 &  & 5  & 41.0  \\
 &  & 10&   45.6 \\ \cmidrule{2-4} 
 & \multicolumn{1}{l|}{\multirow{3}{*}{\textbf{Max Incorrect}}} 
& 1 &    23.9  \\
 & \multicolumn{1}{l|}{} & 3   & 36.0  \\
 & \multicolumn{1}{l|}{} & 5  & 40.5   \\ \midrule
 
 \multirow{8}{*}{\textbf{\begin{tabular}[c]{@{}c@{}}WordNet\\ Similarity\end{tabular}}} &\multirow{4}{*}{\textbf{Max Answers}} 
 & 1   &  45.2  \\
 &  & 3  & 47.8  \\
 &  & 5 & 50.7   \\
 &  & 10 & 55.3 \\ \cmidrule{2-4} 
 & \multicolumn{1}{l|}{\multirow{3}{*}{\textbf{Max Incorrect}}} 
& 1  & 29.2 \\
 & \multicolumn{1}{l|}{} & 3 & 44.6 \\
 & \multicolumn{1}{l|}{} & 5 &  50.6 \\ 
 
 \midrule
 
 \multirow{8}{*}{\textbf{\begin{tabular}[c]{@{}c@{}}RoBERTa\\ Similarity\end{tabular}}}  
 & \multirow{4}{*}{\textbf{Max Answers}} 
 & 1 & 59.0  \\
 &  & 3  & 64.0  \\
 &  & 5  & 66.2   \\
 &  & 10  & 71.7   \\ \cmidrule{2-4} 
 & \multicolumn{1}{l|}{\multirow{3}{*}{\textbf{Max Incorrect}}} 
& 1 & 59.0 \\
 & \multicolumn{1}{l|}{} & 3 &  64.0     \\
 & \multicolumn{1}{l|}{} & 5 & 66.2  \\ 
 \bottomrule
\end{tabular}
 \caption{Results for the ``single human'' ranking scores, replaced by a human evaluation closer to actual task}
\label{table:test-results-backup}
\end{table}

 The human performance numbers reported in \S~4.3 were collected to be maximally similar to the proposed task: like both the training data and the crowdsourced evaluation data, they were generated by asking many humans for a single best answer.  We also collected sets of answers from a small set of in-person annotators using a slightly different questioning paradigm, providing a prompt and asking a single annotator to provide eight different answers to that question.  In practice, we found that this shift in evaluation this could penalize human performance.  One primary issue with this was that the human annotator asked for all answers to the same question would generally only provide a single answer string corresponding to the top answer clusters.  This means that even if the human matched the correct answer,  they would miss that answer cluster entirely if they provided a novel string for that answer cluster.  Annotators also found it be to be quite difficult to provide many answers for the same prompt and would go far afield with later answers, making such answers differ from the distribution of answers in the train and evaluation set.  To avoid confusion using these noticeably different human performance scores, we shifted reporting to a set of data that is closer to the actual task evaluation but report those ranking scores here for transparency. One can see from Table \ref{tab:hrank} and \ref{table:test-results-backup} that such human answers look good, but that the actual scores are dramatically lower than what is seen when humans are evaluated on the same task as the evaluation set, and only barely outperforms a fine-tuned \gpt system.

\end{document}